\documentclass[sigconf]{acmart}

\usepackage{multirow}
\usepackage{booktabs}
\usepackage{colortbl}
\usepackage{xcolor}
\usepackage{siunitx}
\usepackage{float}
\usepackage{stfloats}
\usepackage{placeins}
\usepackage[most]{tcolorbox}
\usepackage{graphicx}
\usepackage{subcaption}

\definecolor{BestLocal}{HTML}{edf7fb} % light green
\definecolor{BestGlobal}{HTML}{C6EFCE} % light blue

\newcommand{\bestbox}[1]{%
  \fcolorbox{purple!60}{purple!10}{\textbf{#1}}%
}

\tcbset{on line,
  boxsep=0pt,left=2pt,right=2pt,top=0pt,bottom=0pt,
  colframe=purple!60!black,
  colback=purple!10,
  rounded corners,
  box align=center,
  nobeforeafter
}

\AtBeginDocument{%
  }

\copyrightyear{2025}
\acmYear{2025}
\setcopyright{cc}
\setcctype{by}
\acmConference[CIIS 2025]{2025 The 8th International Conference on Computational Intelligence and Intelligent Systems}{November 21--23, 2025}{Okayama, Japan}
\acmBooktitle{2025 The 8th International Conference on Computational Intelligence and Intelligent Systems (CIIS 2025), November 21--23, 2025, Okayama, Japan}
\acmPrice{}
\acmDOI{10.1145/3787256.3787261}
\acmISBN{979-8-4007-1933-2/2025/11}

\begin{document}

\title{From Unity Simulation to Diffusion-Based Augmentation: Quantifying Dataset Balance for Robust Object Detection}

\author{Mohamed Benkedadra}
% \email{mohamed.benkedadra@umons.ac.be}
\orcid{0000-0002-2590-8344}
\affiliation{%
  \institution{ILIA, Université de Mons}
  \city{Mons}
  \country{Belgium}
}
% \additionalaffiliation{%
%   \institution{ISIA Lab, University of Mons}
%   \city{Mons}
%   \country{Belgium}
% }
\email{mohamed.benkedadra@umons.ac.be}
\authornote{Corresponding author.}

\author{Aissa Saoudi}
\affiliation{%
  \institution{Université Polytechnique Hauts-de-France}
  \city{Valenciennes}
  \country{France}}
\email{aissa.saoudi@uphf.fr}

\author{Maxime Gloesener}
\affiliation{%
  \institution{DeepILIA, ILIA, Université de Mons}
  \city{Mons}
  \country{Belgium}}
\email{maxime.gloesener@umons.ac.be}

\author{Sidi Ahmed Mahmoudi}
\affiliation{%
  \institution{DeepILIA, ILIA, Université de Mons}
  \city{Mons}
  \country{Belgium}}
\email{sidi.mahmoudi@umons.ac.be}
\orcid{0000-0002-1530-9524}

\author{Matei Mancas}
\affiliation{%
  \institution{ISIA Lab, Université de Mons}
  \city{Mons}
  \country{Belgium}}
\email{matei.mancas@umons.ac.be}
\orcid{0000-0002-6539-6996}

\renewcommand{\shortauthors}{Benkedadra et al.}

\begin{abstract}
Modern computer vision models achieve high accuracy when trained on large-scale annotated datasets. In critical domains such as construction safety monitoring, data collection is costly, hazardous, and ethically constrained. This paper presents a systematic study comparing two complementary data generation paradigms, (1) Unity Simulation-based rendering and (2) Controllable Diffusion-based generation (CIA), for object detection under real data-scarce conditions. A unified experimental framework enables controlled dataset mixing across real, simulated, and generative sources, while maintaining identical model and training settings. Quantitative evaluation using Precision, Recall, mAP, and custom $\Delta$-metrics, reveals that neither simulation nor generative augmentation alone achieves optimal transferability. Unity-only training yields an mAP@0.5 drop of $-50\%$ relative to real data, while CIA-only training shows a milder $-16.5\%$ degradation. Hybrid compositions significantly improve performance, with the 90\% real + 10\% Unity configuration achieving the best overall mAP@0.5 of $62.68\%$ ($+7.64\%$ over baseline), and the 90\% real + 10\% CIA configuration maximizing precision at $74.45\%$. Results demonstrate that limited synthetic inclusion enhances generalization, while excessive substitution induces domain drift.
\end{abstract}


% \keywords{Do, Not, Use, This, Code, Put, the, Correct, Terms, for,


\maketitle

\section{Introduction}

Modern deep object detectors such as YOLO~\cite{redmon2016lookonceunifiedrealtime} achieve remarkable accuracy when trained on large-scale datasets. However, in industrial or safety critical contexts (e.g., construction sites or railway environments) obtaining labeled data is logistically difficult, costly, and often unsafe \cite{Kim2025}. Scenes are dynamic, lighting varies, and safety incidents cannot ethically be staged for data capture. Consequently, these systems are limited by data scarcity rather than by model capacity.

Data scarcity prompted a wide spectrum of strategies aimed at reducing reliance on large annotated datasets. \emph{few-shot}~\cite{wang2020frustratinglysimplefewshotobject, snell2017prototypicalnetworksfewshotlearning} and \emph{zero-shot learning}~\cite{radford2021learningtransferablevisualmodels}, exploit transfer learning and language–vision alignment to generalize from limited examples. \emph{Semi-supervised} and \emph{self-supervised} approaches~\cite{kotthapalli2025self} leverage unlabeled data to learn robust feature representations. \emph{Domain adaptation}~\cite{tzeng2017adversarialdiscriminativedomainadaptation} aligns distributions between source and target domains to mitigate the so called \emph{sim-to-real} gap. 

\begin{enumerate}
    \item \textbf{Simulation-based data generation}, where game/virtual environments platforms render photorealistic and precisely annotated scenes with controllable geometry, lighting, and occlusion patterns. Popular platforms include Unity~\cite{haas2014history, unityUnityRealTime}, Unreal Engine~\cite{unrealengineMostPowerful}, or the more research focused CARLA~\cite{Dosovitskiy17}.

    \item \textbf{Generative data augmentation}, where modern diffusion models such as Stable Diffusion~\cite{rombach2022highresolutionimagesynthesislatent} or ControlNet~\cite{zhang2023addingconditionalcontroltexttoimage} synthesize realistic visual variations conditioned on structure, pose, semantics, etc.
\end{enumerate}

The recently introduced \emph{Controllable Image Augmentation (CIA)} framework~\cite{Benkedadra_2024} unifies and standardizes the second paradigm. This is done by offering modular stages for control condition extraction, conditioned diffusion-based generation, image quality filtering, dataset integration, and parallelized model training. 

Despite their popularity, these paradigms have not been systematically compared under controlled conditions, nor studied jointly in a real-world object detection task. \textbf{This work fills that gap by making the following key contributions:}

\begin{enumerate}
    \item We present a benchmarking between Unity-based simulation and diffusion-based augmentation (CIA) for object detection, under controlled data volume and model settings.

    \item We quantify the separate and combined impact of different real, simulated, and generated data, on detection performance across multiple dataset mixtures. That is done through a joint analysis of simulation domain gap, generative bias, and dataset composition synergy.

    \item We provide a unified, reproducible pipeline built entirely with open-source tools, designed for broader adoption in industrial and academic contexts where synthetic and generative data must be jointly leveraged.
\end{enumerate}

\begin{figure*}[t!]
  \includegraphics[width=\textwidth]{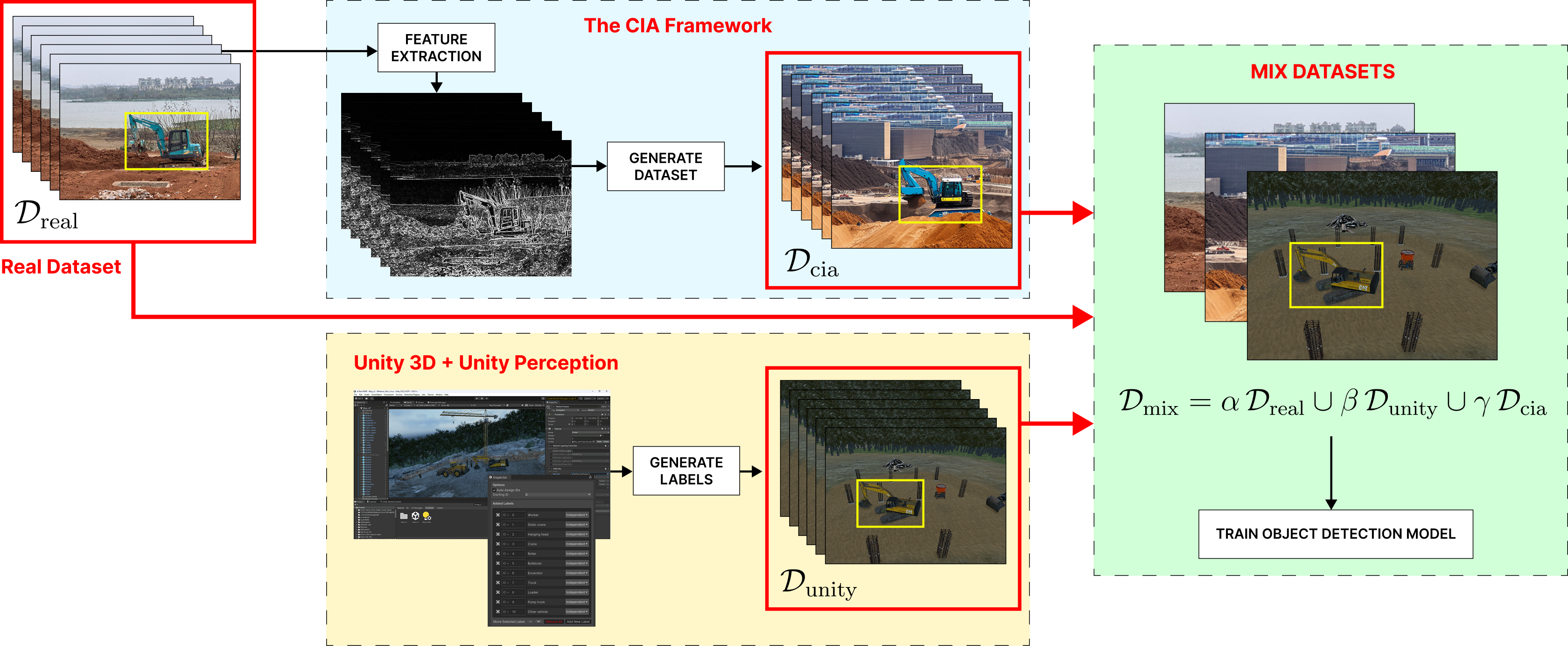}
  \caption{Overview of the proposed hybrid data pipeline. CIA generates photorealistic augmentations from real images using Diffusion, while Unity3D renders synthetic scenes with automatic annotations. The fused dataset $\mathcal{D}_{\text{mix}}$ balances realism and diversity, reducing the \emph{sim-to-real} gap, and improving model robustness.}
  \label{fig:pipeline}
\end{figure*}

\section{Related Work}

Synthetic rendering has been adopted for vision tasks training, due to its controllability, cost-efficiency, and ability to simulate rare objects or scenarios. CARLA for example, published by Dosovitskiy et al. (2014) \cite{Dosovitskiy17}, enabled simulation based autonomous vehicle training and research when real world data capture was dangerous. Zhang et al. (2016) \cite{Zhang2017} systematically studied synthetic rendering for indoor scene understanding, generating 500K synthetic images from 45K realistic 3D scenes. They showed that more realistic rendering improves performance on tasks like semantic segmentation and surface normal prediction.

Many game engines such as Unity3D, offer the ability to develop community plugins. In addition to providing an asset stores where plugins ca be published and sold. Borkman et al. (2021) \cite{borkman2021unityperceptiongeneratesynthetic} published the Unity Perception plugin, which provides a modular framework for large-scale synthetic dataset generation with automatic ground-truth annotations. It is able to generate object bounding boxes, instance segmentation masks, depth maps, and keypoints. The toolkit integrates domain randomization capabilities.

Domain randomization~\cite{tobin2017domainrandomizationtransferringdeep} improves transferability by deliberately varying visual factors such as lighting, textures, object colors, camera poses, and backgrounds during rendering. It forces models to rely on invariant structural cues rather than low-level appearance statistics. Hence, this strategy reduces overfitting to specific synthetic textures and narrows the sim-to-real gap.

Despite these advances, purely randomized rendering often fails to reproduce the complex photometric and material properties of real environments. For example, global illumination effects, fine-grained surface roughness, or realistic motion blur. As a result, models trained solely on domain randomized synthetic data typically under-perform when evaluated on real world imagery~\cite{Asselin2020}. This limitation motivates hybrid approaches that integrate physically grounded simulation with data-driven generative augmentation, as explored in this work through the combination of Unity-based rendering and diffusion-based synthesis.

Diffusion models~\cite{rombach2022highresolutionimagesynthesislatent, ho2020denoisingdiffusionprobabilisticmodels} and ControlNet~\cite{zhang2023addingconditionalcontroltexttoimage} have revolutionized controllable image generation, enabling generative conditioning on edges, poses, or semantic maps. Thus, allowing task specific synthesis. CIA~\cite{Benkedadra_2024} leverages diffusion models for data augmentation via extraction of structural features and conditions, controlled synthesis, and quality filtering. This type of augmentation has proven effective, but pretrained diffusion models can introduce semantic bias from internet scale training corpora~\cite{yamaguchi2023limitationdiffusionmodelssynthesizing, luccioni2023stablebiasanalyzingsocietal, perera2023analyzingbiasdiffusionbasedface}.

Bridging the sim-to-real gap has been a longstanding challenge~\cite{tobin2017domainrandomizationtransferringdeep, Peng_2018}. Recent data-centric AI paradigms~\cite{jakubik2024datacentricartificialintelligence} emphasize dataset quality, balance, and representativeness over model complexity. Our work aligns with this direction by empirically quantifying how synthetic and generative data affect generalization in real-world evaluation.

\section{Methodology}

We aim to systematically analyze how synthetic simulation and generative diffusion, influence object detection robustness under data-scarce industrial scenarios.

The proposed pipeline, illustrated in Figure~\ref{fig:pipeline}, is divided into four components. (1) Controllable Diffusion-based (CIA) data generation, (2) Unity Simulation-based data generation, (3) dataset composition and balancing, and (4) Model training and evaluation.

\subsection{Data Baseline and Augmentation Strategies}

Let $\mathcal{D}_{\text{real}}$ denote the real baseline dataset, consisting of RGB images $I_i$, and their corresponding labels $y_i$ (object class and bounding box annotations). This dataset represents the operational target domain but is inherently limited in scale and variability.
\begin{equation}
    \mathcal{D}_{\text{real}} = \{(I_i, y_i)\}_{i=1}^{N}
\end{equation}

\textbf{In Diffusion-based augmentation,} the Controllable Image Augmentation (CIA) framework~\cite{Benkedadra_2024} enhances data diversity by generating photorealistic variants of existing samples while preserving their structural semantics.

Given an image $I_i \in \mathcal{D}_{\text{real}}$, a configurable encoder $\mathcal{E}$ extracts structural control features. $F_i$ can represent edge maps, depth projections, or semantic segmentation masks depending on the selected control mode.
\begin{equation}
F_i = \mathcal{E}(I_i),
\label{eq:cia_encoder}
\end{equation}

A controlled diffusion generator $\mathcal{G}$ (e.g., Stable Diffusion~\cite{rombach2022highresolutionimagesynthesislatent} augmented with ControlNet~\cite{zhang2023addingconditionalcontroltexttoimage}) is conditioned on $F_i$ and a textual prompt $C$ to synthesize a new image $\tilde{I}_i$.
\begin{equation}
\tilde{I}_i = \mathcal{G}(F_i, C).
\label{eq:cia_generation}
\end{equation}

The generated image $\tilde{I}_i$ inherits the annotation $y_i$ from its real counterpart $I_i$, yielding the diffusion-augmented dataset.
\begin{equation}
\mathcal{D}_{\text{cia}} = \{(\tilde{I}_i, y_i)\}_{i=1}^{N}.
\end{equation}

To ensure photometric realism and semantic fidelity, each generated image is evaluated through a composite \emph{quality control function} $q(\tilde{I}_i)$, where different metrics like $\text{FID}$, $\text{NIMA}$ and $\text{CLIP}$ ensure high-quality, label-consistent augmentation.

This process results in a generative dataset $\mathcal{D}_{\text{cia}}$ that complements $\mathcal{D}_{\text{real}}$ by expanding its visual diversity while maintaining semantic alignment with real-world conditions.

\textbf{In Unity-based simulation augmentation}, synthetic datasets are procedurally generated using the Unity Perception toolkit~\cite{borkman2021unityperceptiongeneratesynthetic}. Each sample is produced by rendering a fully controllable 3D environment with randomized parameters governing lighting, camera pose, surface materials, and object placement, following the domain randomization principle~\cite{tobin2017domainrandomizationtransferringdeep}.  
This approach enables systematic exploration of geometric and photometric variations that are difficult or unsafe to capture in the real world. Formally, the simulated dataset is defined as ${D}_{\text{unity}}$ where each $I_j^{u}$ represents a rendered RGB image and $y_j^{u}$ its corresponding ground-truth label set (object bounding boxes, segmentation masks, and class identifiers). All annotations are automatically generated at render time through the Unity Perception API, eliminating manual labeling cost and errors.
\begin{equation}
    \mathcal{D}_{\text{unity}} = \{(I_j^{u}, y_j^{u})\}_{j=1}^{N}
\end{equation}

Unity provides fine-grained control over environmental parameters $\theta_u = \{$illumination, materials, occlusion, 
camera pose, scene layout, etc$\}$ allowing domain randomization across a high dimensional visual parameter space. By varying $\theta_u$ across render iterations, the simulator yields a diverse, densely annotated dataset that enhances geometric and contextual coverage. However, despite this controllability, synthetic images often diverge from real-world visual statistics. This is due to limited photometric realism. For example, global illumination, micro-texture, and noise characteristics are hard to reproduce.

This discrepancy constitutes the well-known \emph{simulation-to-reality (sim-to-real) gap}, which can impair generalization when models trained on $\mathcal{D}_{\text{unity}}$ are evaluated on $\mathcal{D}_{\text{real}}$.  To mitigate this gap, the subsequent CIA-based generative augmentation stage injects learned visual priors from real imagery. Hence, blending the strengths of procedural-based simulation with the strengths of diffusion-based realism.

\textbf{To compose balanced datasets mixes,} we consider a real dataset $\mathcal{D}_{\text{real}}$ of $N$ labeled samples. We also consider a Unity-rendered dataset $\mathcal{D}_{\text{unity}}$ and CIA-augmented dataset $\mathcal{D}_{\text{cia}}$. Since each CIA-generated image $\tilde{I}_i$ originates from a real sample $I_i$ through a one-to-one generation strategy (Eq.~\ref{eq:cia_generation}), all datasets should share identical cardinality.
\begin{equation}
|\mathcal{D}_{\text{real}}| = |\mathcal{D}_{\text{unity}}| = |\mathcal{D}_{\text{cia}}| = N.
\label{eq:dataset_equal_size}
\end{equation}

To study how the substitution of real images with synthetic or generative ones affects model robustness, we construct a mixed dataset $\mathcal{D}_{\text{mix}}$ by replacing a proportion of $\mathcal{D}_{\text{real}}$ with samples from $\mathcal{D}_{\text{unity}}$ and $\mathcal{D}_{\text{cia}}$.  Formally, the dataset mixture is parameterized by ratios $(\alpha, \beta, \gamma)$ such that :
\begin{equation}
\mathcal{D}_{\text{mix}}(\alpha, \beta, \gamma)
= \alpha\, \mathcal{D}_{\text{real}}
  \cup \beta\, \mathcal{D}_{\text{unity}}
  \cup \gamma\, \mathcal{D}_{\text{cia}},
\quad
\alpha + \beta + \gamma = 1.
\label{eq:dataset_mix_equal_size}
\end{equation}

This formulation enables isolating the individual and joint effects of simulation and diffusion on generalization, under identical training volumes. In practice, we evaluate configurations such as $(1,0,0)$, $(0,1,0)$, $(0,0,1)$, and mixes like $(0.8,0.15,0.05)$ to quantify trade-offs between realism, diversity, and performance.

\subsection{Model Training and Evaluation Metrics}

All experiments employ a consistent detection architecture $\mathcal{M}_{\phi}$ trained under identical optimization and scheduling conditions to ensure that performance variations stem solely from data composition. All trained models are evaluated on a held-out real-world test set $\mathcal{D}_{\text{real}}^{\text{test}}$, ensuring that no synthetic or generative images are seen during evaluation. Performance is reported in terms of mean Average Precision (mAP), Precision, and Recall. For consistency, all metrics are computed at IoU thresholds of 0.5 and 0.5:0.95 following COCO evaluation standards.

To quantify the influence of dataset origin on detection robustness, we define two comparative metrics that characterize the relative performance gap between real, simulated, and generative data~:
\begingroup
\setlength{\abovedisplayskip}{5pt}
\setlength{\belowdisplayskip}{5pt}
\begin{align}
\Delta_{\text{sim→real}}(\alpha,\beta) &=
\text{mAP}(\mathcal{D}_{\text{mix}}(\alpha,\beta,0))
- \text{mAP}(\mathcal{D}_{\text{real}}), \label{eq:delta_sim_ratio}\\[3pt]
\Delta_{\text{gen→real}}(\alpha,\gamma) &=
\text{mAP}(\mathcal{D}_{\text{mix}}(\alpha,0,\gamma))
- \text{mAP}(\mathcal{D}_{\text{real}})
\label{eq:delta_gen_ratio}
\end{align}
\endgroup

$\Delta_{\text{sim→real}}$ quantifies the simulation domain gap, reflecting the loss in realism induced by photometric, geometric, and material discrepancies, between Unity-rendered and real images. $\Delta_{\text{gen→real}}$ captures the generative bias introduced by Diffusion-based augmentation, encompassing both the visual realism and the semantic priors inherited from pretrained generative models. 

Together, these ratio-dependent metrics form a rigorous framework for assessing how simulation and generative synthesis contribute to real-world object detection robustness.  

In summary, our methodology provides a reproducible protocol for isolating and quantifying the effects of data origin. By controlling data mixture ratios, and maintaining identical training pipelines, we explicitly measure the trade-off between \emph{realism}, \emph{controllability}, and \emph{semantic bias}. Thus, yielding actionable insights for data centric model design in computer vision applications.

\section{Experimental Setup}

All experiments are conducted on the MOCS dataset~\cite{XUEHUI2021103482}, a construction site dataset designed for safety and activity monitoring. It contains diverse outdoor scenes with workers, helmets, machinery, and equipment. These scenes were captured under varying illumination, occlusion, and weather conditions. Each image is annotated with bounding boxes and class identifiers corresponding to safety related entities (e.g. helmet). The real subset $\mathcal{D}_{\text{real}}$ used in this study contains $N=3000$ labeled images sampled to maximize contextual diversity across recording conditions.

To complement real data, a synthetic dataset $\mathcal{D}_{\text{unity}}$ was generated using the Unity game engine. The simulated environment replicates a construction site populated with human avatars, cranes, and vehicles, rendered with physically based textures \cite{PharrPhysicallyBasedTexture}. Randomization was applied to environmental parameters $\theta_u$. Each rendered frame automatically includes 2D bounding boxes, instance masks, and class annotations exported by the Unity Perception API. The resulting dataset maintains parity in size with $\mathcal{D}_{\text{real}}$ to ensure balanced comparison.

The generative dataset $\mathcal{D}_{\text{cia}}$ was produced using the Controllable Image Augmentation (CIA) framework using the canny edge extraction method, which was shown to produce the most improvement in model performance, out of all the extractor-generator couples introduced in the original paper. 

\begin{figure}[h]
    \centering
    \includegraphics[width=\linewidth]{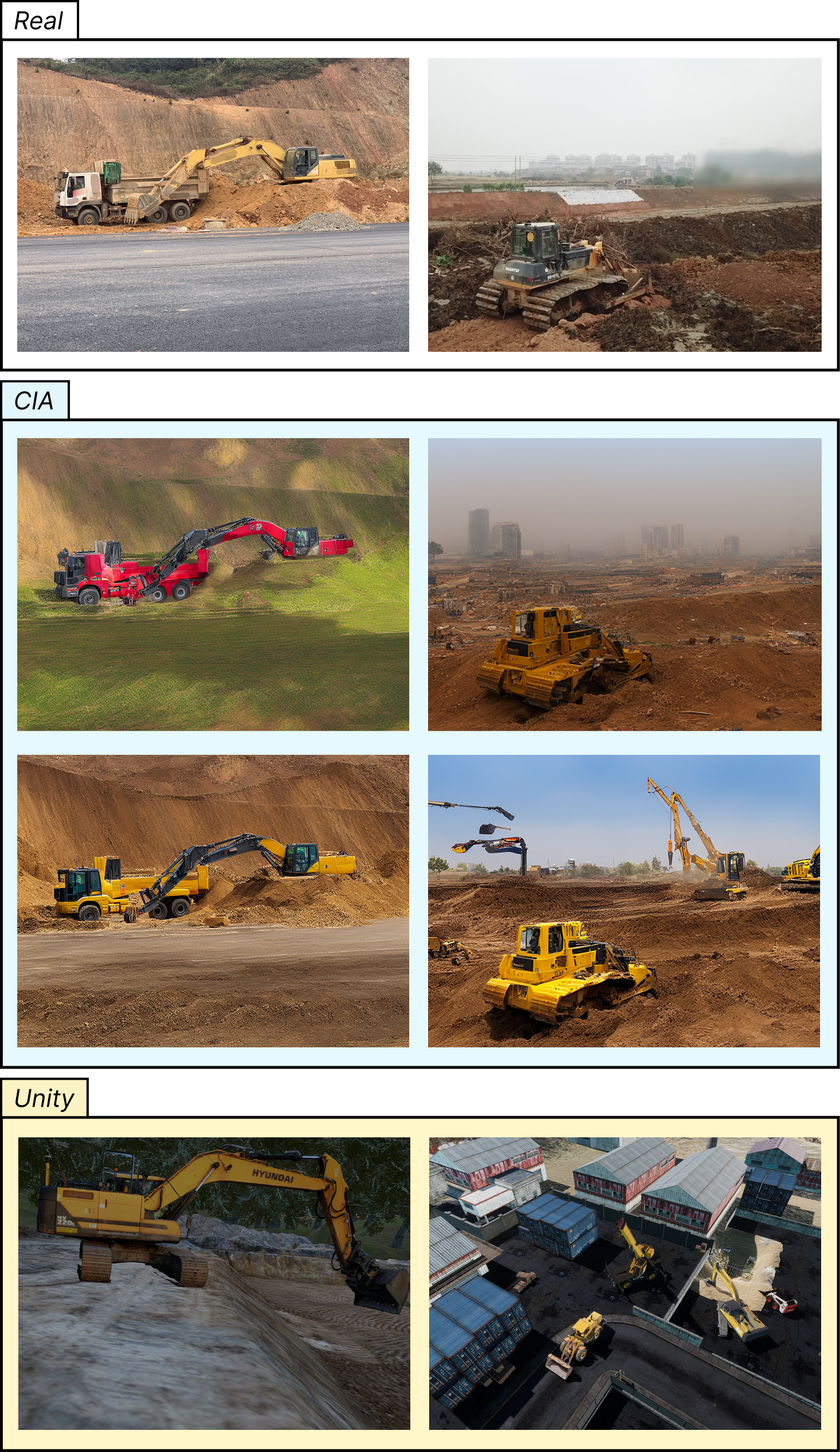}
    \caption{Representative samples from the three dataset sources used in this study. Top: real construction-site images from MOCS. Middle: CIA-generated variants using diffusion-based controllable augmentation. Bottom: Unity-rendered synthetic scenes produced through domain randomization.}
    \label{fig:dataset_samples}
\end{figure}

As a result, we produced three datasets where $|\mathcal{D}_{\text{real}}| = |\mathcal{D}_{\text{unity}}| = |\mathcal{D}_{\text{cia}}| = N$. Qualitative examples of Unity-rendered, CIA-generated, and real-world samples are shown in Figure~\ref{fig:dataset_samples}. We used a test dataset of 300 real images for evaluation.

Training and evaluation were performed using YOLOv11n\footnote{The Ultralytics~v8.3 implementation of YOLOv11 was used for experimentation}. All experiments ran on Google Colab Pro environments with NVIDIA L4 GPUs, Python~3.10, and PyTorch~2.3. Each model was trained for 20~epochs with a batch size of 64, image resolution of 640$\times$640. Adam was used for optimization, with a learning rate $\alpha=1\times10^{-3}$ and cosine decay scheduling. Standard data augmentations (flips, scaling, hue-saturation jitter) were applied identically across all runs.

For each experiment, a mixed dataset $\mathcal{D}_{\text{mix}}$ was constructed, varying the ratios of real, synthetic, and generative data. Evaluated configurations included both pure and hybrid compositions such as $(1,0,0)$, $(0,1,0)$, $(0,0,1)$, $(0.75,0.25,0)$, $(0.33,0.33,0.33)$, etc. All models were tested on a held-out real validation set $\mathcal{D}_{\text{real}}^{\text{test}}$ unseen during training.

Performance is reported using mean Average Precision (mAP) at IoU thresholds 0.5 (mAP@0.5) and 0.5:0.95 (mAP@0.5:0.95), along with Precision, Recall, and Fitness. To quantify the influence of data composition, the metrics $\Delta_{\text{sim→real}}$ and $\Delta_{\text{gen→real}}$ (Eqs.~\ref{eq:delta_sim_ratio}–\ref{eq:delta_gen_ratio}) are computed for all configurations, capturing how simulation and diffusion influence generalization performance on real-world test data.

\section{Results}

\begin{table*}[bp]
\centering
\caption{
Comprehensive performance comparison across all dataset composition experiments. Each configuration reports Precision, Recall, mAP@0.5, mAP@0.5:0.95, and Fitness, evaluated on a held-out real test set. Groups correspond to different mixing regimes between Real, Unity-simulated, and CIA-generated data. Green-highlighted row indicates the best overall result, while Blue-highlighted rows indicate the best results within each group. Purple-bordered cells mark the overall best value per metric, across all configurations.}
\label{tab:grouped_results_highlighted}
\renewcommand{\arraystretch}{1.15}
\setlength{\tabcolsep}{5pt}
\rowcolors{2}{gray!8}{white}
\begin{tabular}{
l
S[table-format=3.0]
S[table-format=3.0]
S[table-format=3.0]
S[table-format=1.3]
S[table-format=1.4]
S[table-format=1.4]
S[table-format=1.4]
S[table-format=1.4]
}
\toprule
\textbf{Group} & \textbf{Real} & \textbf{CIA} & \textbf{Unity} & 
\textbf{Precision} & \textbf{Recall} & \textbf{mAP@0.5} & 
\textbf{mAP@0.5:0.95} & \textbf{Fitness} \\
\midrule

\rowcolor{gray!20}
\textbf{Baselines} & & & & & & & & \\
Real only & 100 & 0 & 0 & \cellcolor{BestLocal}\textbf{0.7266} & \cellcolor{BestLocal}\textbf{0.4828} & \cellcolor{BestLocal}\textbf{0.5504} & \cellcolor{BestLocal}\textbf{0.3873} & \cellcolor{BestLocal}\textbf{0.4036} \\
CIA only & 0 & 100 & 0 & 0.5923 & 0.3566 & 0.3851 & 0.2551 & 0.2681 \\
Unity only & 0 & 0 & 100 & 0.3004 & 0.0558 & 0.0447 & 0.0260 & 0.0279 \\

\rowcolor{gray!20}
\textbf{Real--CIA Mixes} & & & & & & & & \\
 & 90 & 10 & 0 & \cellcolor{BestLocal}\bestbox{0.7445} & \cellcolor{BestLocal}\textbf{0.5063} & \cellcolor{BestLocal}\textbf{0.5792} & \cellcolor{BestLocal}\textbf{0.4076} & \cellcolor{BestLocal}\textbf{0.4248} \\
 & 75 & 25 & 0 & 0.6530 & 0.4591 & 0.4998 & 0.3345 & 0.3582 \\
 & 50 & 50 & 0 & 0.6543 & 0.4449 & 0.4940 & 0.3427 & 0.3578 \\
 & 25 & 75 & 0 & 0.6518 & 0.4584 & 0.4965 & 0.3342 & 0.3504 \\
 & 10 & 90 & 0 & 0.6423 & 0.4118 & 0.4567 & 0.3088 & 0.3236 \\

\rowcolor{gray!20}
\textbf{Real--Unity Mixes} & & & & & & & & \\
 & 90 & 0 & 10 & \cellcolor{BestGlobal}\textbf{0.7400} & \cellcolor{BestGlobal}\bestbox{0.5739} & \cellcolor{BestGlobal}\bestbox{0.6268} & \cellcolor{BestGlobal}\bestbox{0.4481} & \cellcolor{BestGlobal}\bestbox{0.4660} \\
 & 75 & 0 & 25 & 0.5551 & 0.3811 & 0.4079 & 0.2766 & 0.2898 \\
 & 50 & 0 & 50 & 0.4876 & 0.3392 & 0.3354 & 0.2196 & 0.2312 \\
 & 25 & 0 & 75 & 0.3528 & 0.2206 & 0.1993 & 0.1214 & 0.1293 \\
 & 10 & 0 & 90 & 0.3527 & 0.2205 & 0.1993 & 0.1215 & 0.1293 \\

\rowcolor{gray!20}
\textbf{Tri-Source Mixes} & & & & & & & & \\
 & 50 & 25 & 25 & \cellcolor{BestLocal}\textbf{0.6192} & \cellcolor{BestLocal}\textbf{0.4362} & \cellcolor{BestLocal}\textbf{0.4712} & \cellcolor{BestLocal}\textbf{0.3229} & \cellcolor{BestLocal}\textbf{0.3377} \\
 & 33 & 33 & 33 & 0.6166 & 0.4333 & 0.4665 & 0.3201 & 0.3347 \\
 & 25 & 25 & 50 & 0.5997 & 0.3890 & 0.4282 & 0.2905 & 0.3043 \\

\rowcolor{gray!20}
\textbf{CIA--Unity Mixes} & & & & & & & & \\
 & 0 & 90 & 10 & \cellcolor{BestLocal}\textbf{0.5395} & \cellcolor{BestLocal}\textbf{0.3523} & \cellcolor{BestLocal}\textbf{0.3678} & \cellcolor{BestLocal}\textbf{0.2455} & \cellcolor{BestLocal}\textbf{0.2577} \\
 & 0 & 75 & 25 & 0.5202 & 0.3326 & 0.3569 & 0.2408 & 0.2524 \\
 & 0 & 50 & 50 & 0.5173 & 0.3141 & 0.3324 & 0.2217 & 0.2328 \\
 & 0 & 25 & 75 & 0.4385 & 0.2693 & 0.2705 & 0.1737 & 0.1834 \\
 & 0 & 10 & 90 & 0.3081 & 0.2087 & 0.1802 & 0.1086 & 0.1158 \\

\bottomrule
\end{tabular}
\end{table*}

Table~\ref{tab:grouped_results_highlighted} reports the detection performance across all dataset compositions. The 100\% real configuration $(1,0,0)$ serves as the baseline. CIA-only $(0,1,0)$ and Unity-only $(0,0,1)$ variants isolate the effects of generative and simulated data respectively. Intermediate mixtures quantify how controlled substitution of real samples affects generalization when evaluated on $\mathcal{D}_{\text{real}}^{\text{test}}$. The results confirm three main observations:

\begin{enumerate}
    \item Simulation-only training suffers from severe domain mismatch, yielding an mAP@0.5 drop of over 50\% compared to the real baseline.
    \item Pure diffusion-based augmentation performs moderately better, but remains below real-only training.
    \item Controlled hybrid ratios, particularly 90\% real with 10\% Unity or CIA, achieve superior generalization, improving up to $+0.076$ mAP@0.5 relative to the baseline.
\end{enumerate}

A closer inspection of Table~\ref{tab:grouped_results_highlighted} reveals a consistent divergence between precision-oriented and recall-oriented trends across the two hybrid regimes. The \emph{Real-CIA mixtures} exhibit the highest precision values, peaking at 0.7445 for the 90\%/10\% configuration. That is slightly above the 100\% Real baseline at 0.7266. This indicates that diffusion-based augmentation enhances the detector's ability to produce low False-Positive predictions. 

Conversely, the \emph{Real-Unity mixtures} outperform all other groups in recall, mAP@0.5, mAP@0.5:0.95, and Fitness, reaching their peak at the 90\%/10\% configuration. Overall, these complementary effects underline the trade-off between \emph{precision-oriented realism} (from CIA) and \emph{recall-oriented diversity} (from Unity).

\begin{figure*}[t!]
    \centering
    % ================= TOP ROW =================
    \begin{minipage}[t]{0.32\textwidth}
        \centering
        \includegraphics[width=\linewidth]{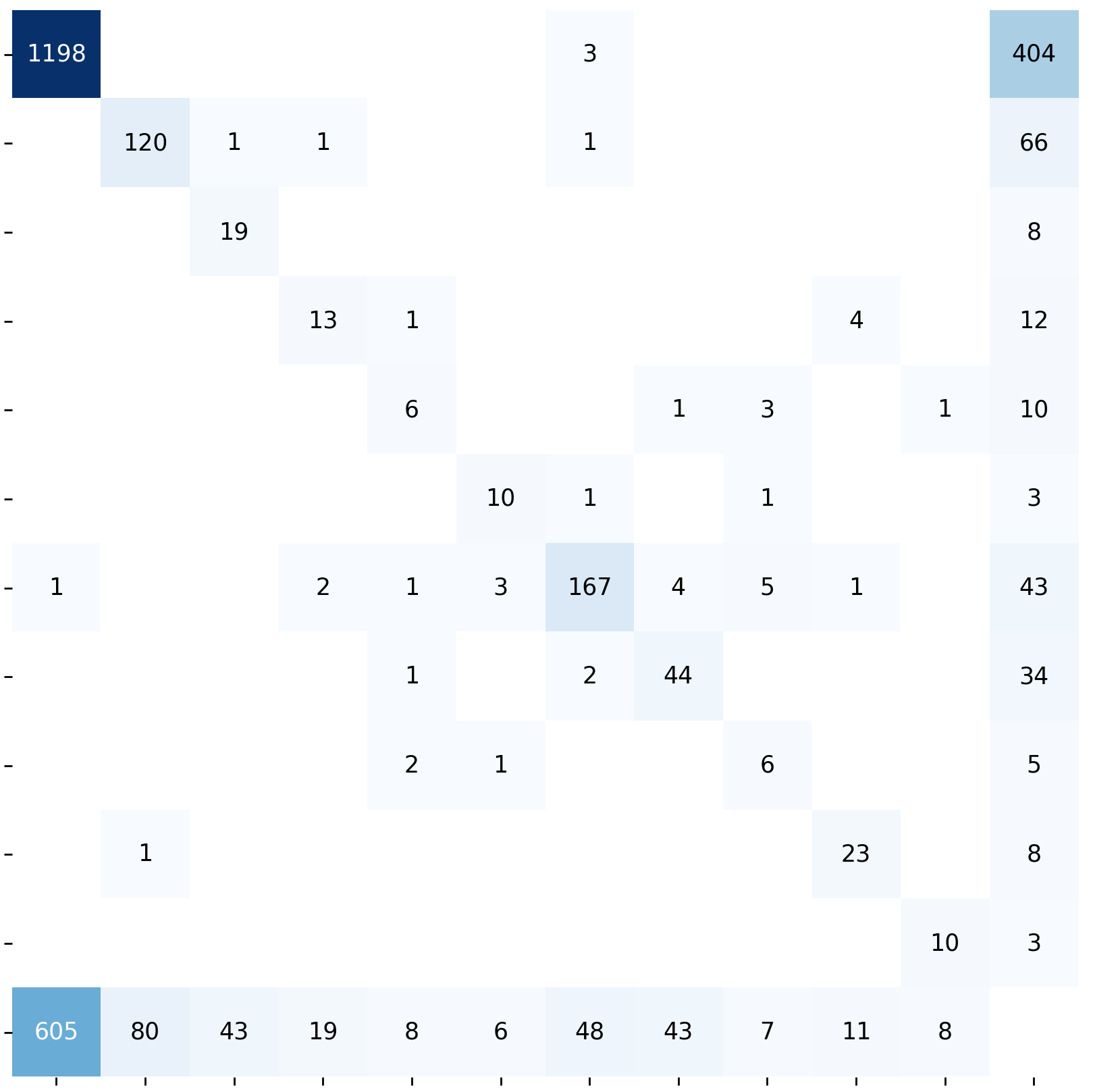}
        \subcaption{100\% Real (MOCS)}
        \label{fig:conf_real100}
    \end{minipage}
    \hfill
    \begin{minipage}[t]{0.32\textwidth}
        \centering
        \includegraphics[width=\linewidth]{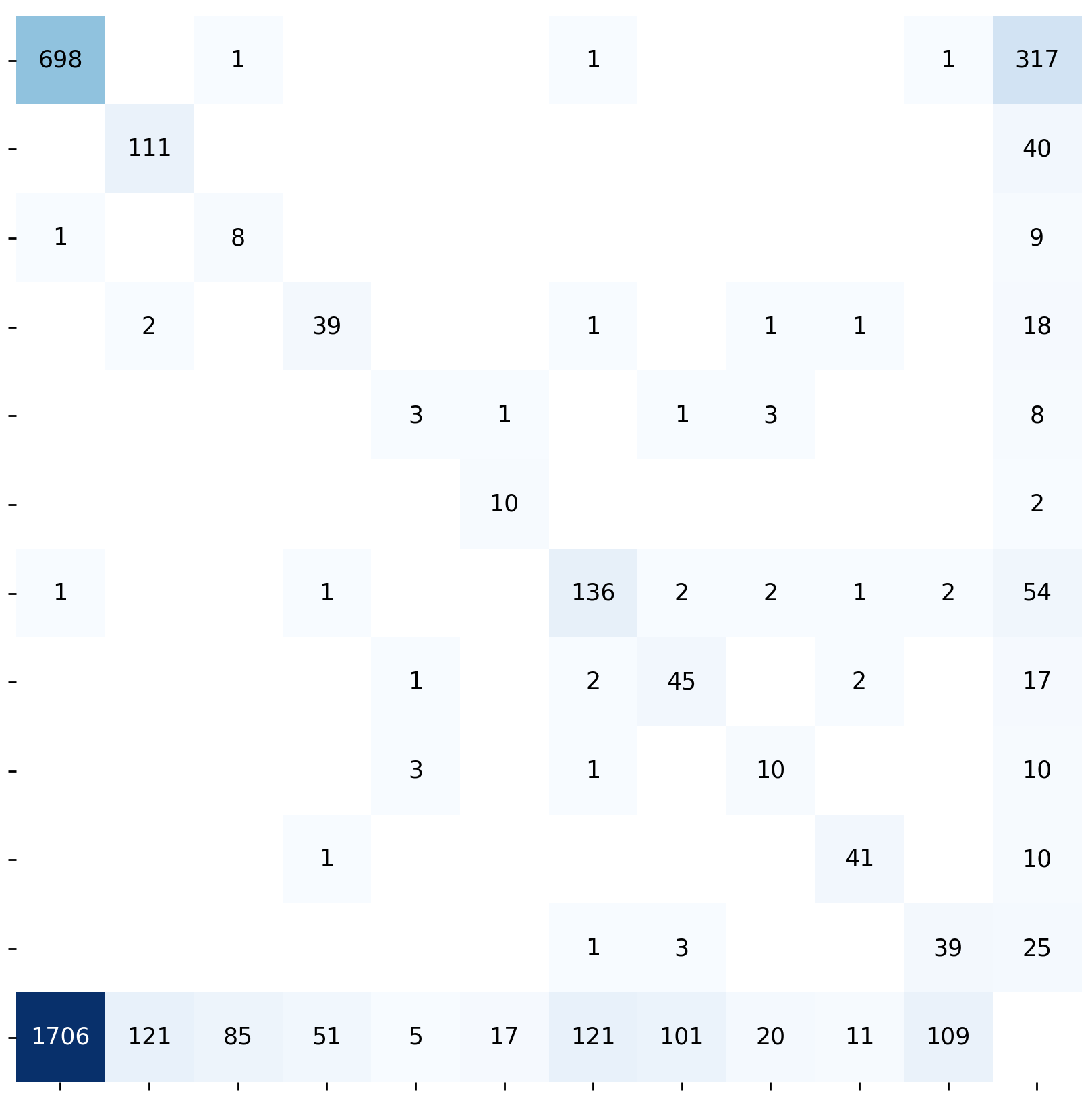}
        \subcaption{100\% CIA}
        \label{fig:conf_cia100}
    \end{minipage}
    \hfill
    \begin{minipage}[t]{0.32\textwidth}
        \centering
        \includegraphics[width=\linewidth]{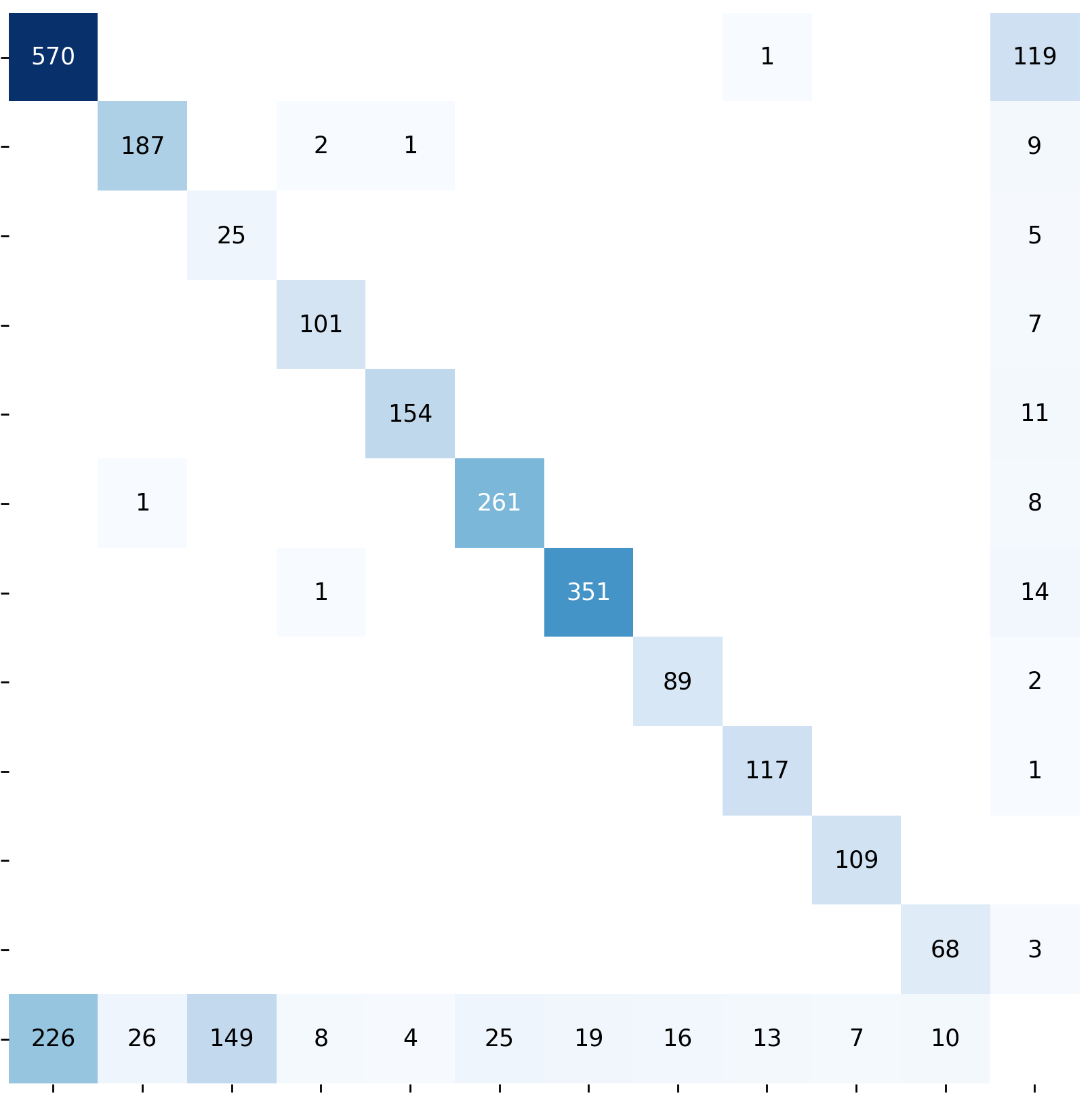}
        \subcaption{100\% Unity}
        \label{fig:conf_unity100}
    \end{minipage}

    \vspace{1em}

    % ================= BOTTOM ROW =================
    \begin{minipage}[t]{0.485\textwidth}
        \centering
        \includegraphics[width=\linewidth]{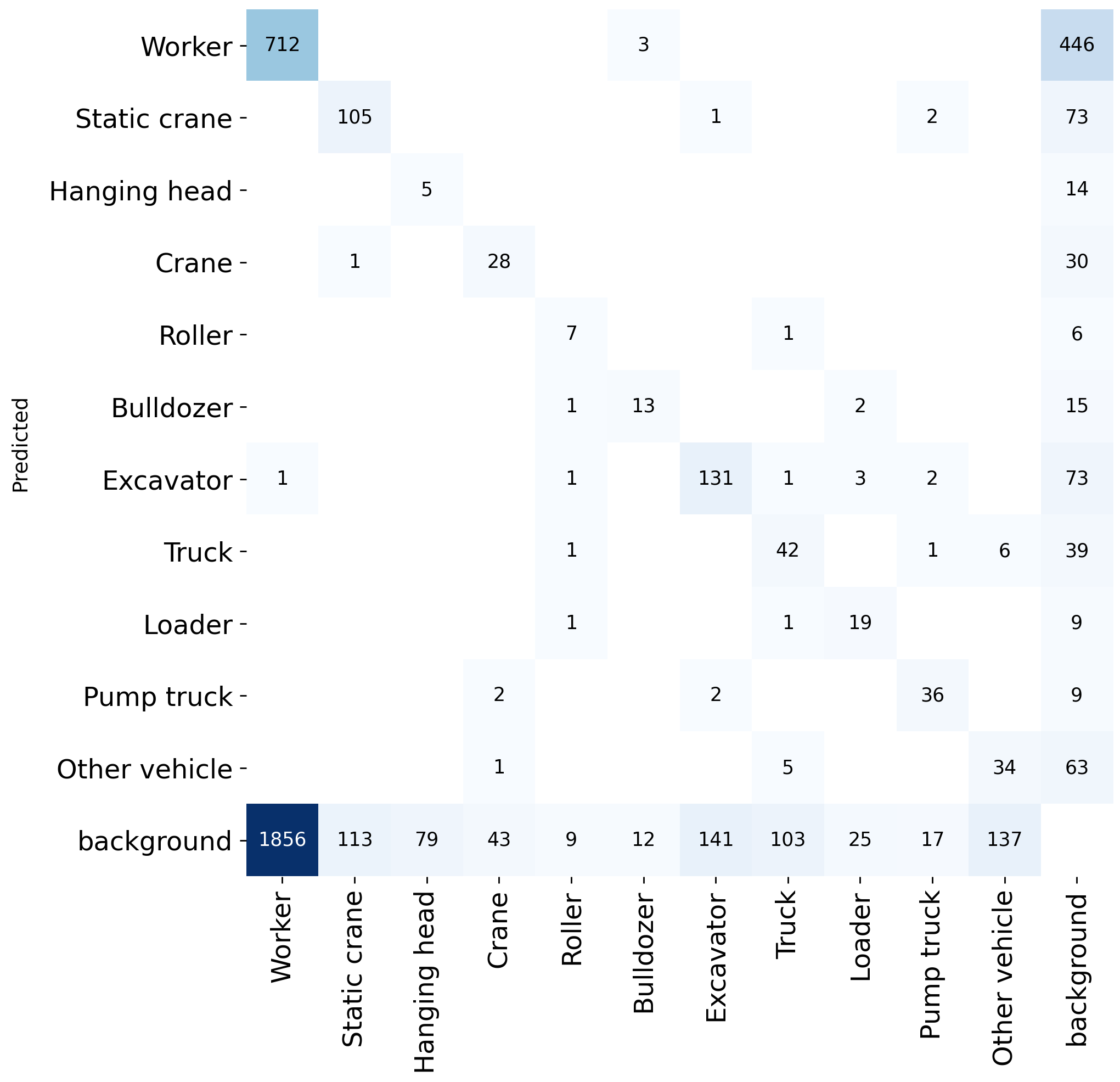}
        \subcaption{90\% Real + 10\% CIA}
        \label{fig:conf_real90_cia10}
    \end{minipage}
    \hfill
    \begin{minipage}[t]{0.46\textwidth}
        \centering
        \includegraphics[width=\linewidth]{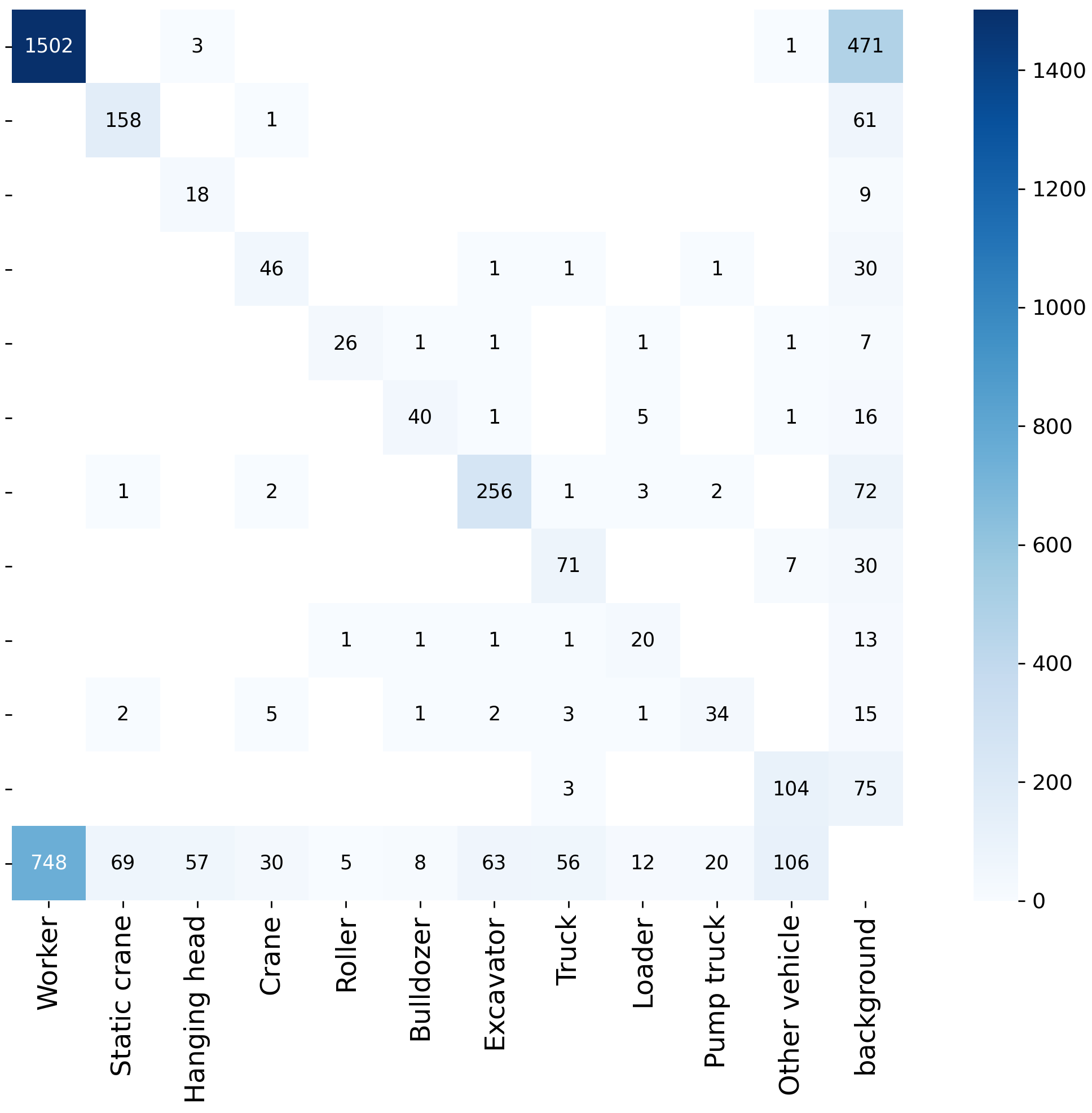}
        \subcaption{90\% Real + 10\% Unity}
        \label{fig:conf_real90_unity10}
    \end{minipage}
    \hfill

    \caption{Non-normalized confusion matrices with a shared color scale. Raw counts expose the dominant error mode (FN→background) and the reduction achieved by small Unity/CIA mixes.}
    \label{fig:conf_matrices_combined}
\end{figure*}

The qualitative results visualized in Figure~\ref{fig:conf_matrices_combined} corroborate the quantitative trends observed in Table~\ref{tab:grouped_results_highlighted}. Because the MOCS dataset is highly imbalanced, we report non-normalized matrices to preserve the true distribution of errors. The 100\% Real baseline (Figure~\ref{fig:conf_real100}) exhibits a strong diagonal across frequent classes (\textit{Worker}, \textit{Static crane}, \textit{Excavator}), with most residual errors corresponding to false negatives assigned to the background. 

In contrast, the 100\% Unity model (Figure~\ref{fig:conf_unity100}) collapses toward background predictions, revealing severe domain shift and explaining its sharp mAP degradation. The 100\% CIA configuration (Figure~\ref{fig:conf_cia100}) retains photometric realism, but shows weaker diagonality on structural and spatially complex classes, reflecting limited geometric variability. Introducing a limited amount of synthetic data restores class-specific diagonality. CIA mixing (Figure~\ref{fig:conf_real90_cia10}) suppresses off-diagonal confusions between visually similar categories. Hence, leading to higher precision. Unity mixing Figure~\ref{fig:conf_real90_unity10} visibly strengthens the main diagonal, by increasing true positives, even though the off-diagonal mass remains.

\begin{figure*}[t!]
    \centering
    \includegraphics[width=0.7\linewidth]{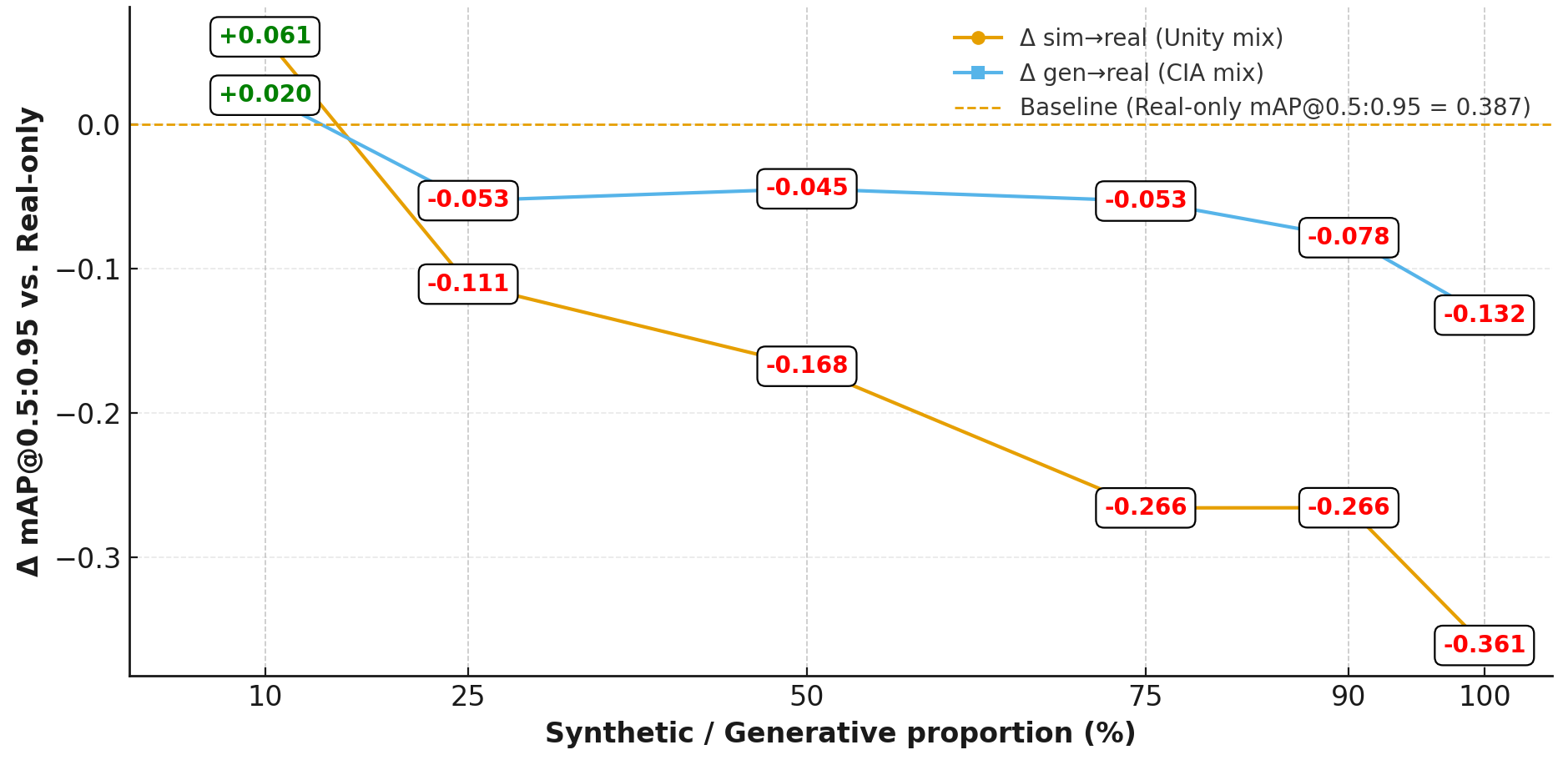}
    \caption{Comparison of $\Delta_{\text{sim→real}}$ and $\Delta_{\text{gen→real}}$ across increasing synthetic data ratios}
    \label{fig:delta_metrics}
\end{figure*}

To quantify domain and generative effects independently of absolute scores, Figure~\ref{fig:delta_metrics} summarizes the relative $\Delta$ metrics as defined in Eqs.~\ref{eq:delta_sim_ratio}–\ref{eq:delta_gen_ratio}. Positive values indicate performance improvement over the real-only baseline, while negative values denote degradation.

We notice that both $\Delta_{\text{sim→real}}$ and $\Delta_{\text{gen→real}}$ follow an inverted-U trend. Moderate substitution (10\%) yields positive gains, while higher proportions cause overfitting. This behavior empirically validates the controlled augmentation hypothesis which states that synthetic data is beneficial only when constrained to limited ratios. the $\Delta_{\text{gen→real}}$ curve decays more slowly than $\Delta_{\text{sim→real}}$, indicating that CIA-based generative data maintains transferability under higher substitution ratios. In contrast, Unity-rendered data exhibits faster degradation, reflecting stronger domain divergence as synthetic content increases.

\FloatBarrier

\section{Discussion}

The results presented in Table~\ref{tab:grouped_results_highlighted} and Figure~\ref{fig:delta_metrics} reveal a nuanced interplay between dataset realism, diversity, and controllability in shaping generalization performance. Neither purely synthetic nor purely generative augmentation alone achieves optimal transferability. 10\% CIA substitution insures the highest precision, while 10\% Unity based substitution insures the highest recall. This precision–recall asymmetry is visually supported by the confusion matrices (Figure~\ref{fig:conf_matrices_combined}).

The degradation observed for the Unity-only configuration (a drop of over 50\% in mAP@0.5:0.95 relative to the real baseline) reflects the well-documented \emph{simulation-to-reality gap}. Despite extensive domain randomization, Unity-rendered scenes diverge from real photometric distributions. Global illumination, specular reflections, and fine-grained textures remain simplified. This induces a representational bias where detectors overfit to synthetic regularities, such as unrealistic sharp edges. Nonetheless, introducing a limited fraction of simulated samples (10\%) substantially improves recall and overall mAP. This improvement indicates that simulated data acts as a diversity regularizer. It plays a role in expanding geometric coverage and reducing overfitting to the narrow appearance manifold of the real subset.

The CIA-only configuration exhibits a milder degradation, confirming that diffusion-based augmentation better preserves real-world statistics. Diffusion models inherit strong natural priors from large-scale image corpora. This knowledge allows for realistic illumination and texture reproduction. Hence, explaining the higher precision in Real-CIA mixtures. Particularly at 90\%/10\% ratios, where the model achieves the lowest recorded false-positive rate. The improvement suggests that CIA reinforces the photometric discriminability of the detector. However, recall remains comparatively limited, implying that generative augmentation introduces less geometric diversity. Diffusion-based synthesis primarily perturbs textures and lighting while maintaining similar spatial layouts, leading to photometric realism but limited structural novelty. Moreover, pretrained diffusion models can embed latent semantic biases from their internet-scale training corpora. Thus, favoring familiar object configurations and backgrounds over rare domain specific contexts.

The superior performance of the 90\% Real + 10\% Unity configuration demonstrates that a small synthetic contribution can enhance generalization without overwhelming the real data distribution. This composition combines Unity's geometric variability with the contextual grounding of real data, offering complementary bias compensation. Yet, as mixing ratios increase, the divergence between simulation and reality grows, ultimately harming convergence and stability. Interestingly, tri-source compositions (e.g., 50/25/25) perform worse than their dual counterparts, suggesting that mixing simulation and generative domains simultaneously can introduce conflicting statistical cues. While Unity images broaden geometric space, CIA images densify photometric space. Excessive blending of both may confuse the model's internal domain boundaries, leading to representational interference. These observations imply that hybrid training could benefit from \emph{progressive curriculum mixing}. This is done starting with simulation-heavy training for structural generalization. Then, gradually transitioning toward CIA and real data for photometric alignment.

Aggregating $\Delta_{\text{sim→real}}$ and $\Delta_{\text{gen→real}}$ across all configurations highlights distinct decay patterns. Generative data maintains positive deltas for broader substitution ratios, whereas simulated data exhibits faster degradation. \textbf{From a bias-variance perspective, Unity samples introduce high-variance perturbations that initially aid generalization, but quickly destabilize learning when the variance exceeds the model's tolerance. CIA on the other hand induces low-variance, low-bias perturbations. Thus, maintaining domain alignment longer, albeit with limited geometric expansion.} This asymmetry underscores the complementary nature of both data sources. Simulation drives diversity, while generative diffusion drives realism. Their joint utility lies in controlled proportionality, not volume.

Several avenues exist to further understand and enhance these effects. Feature-level analysis could reveal which network layers benefit most from each modality. As a result, distinguishing whether improvements occur at low-level edge encoding or high-level semantic abstraction. Evaluating models on out-of-domain real datasets (e.g., construction sites under novel lighting, culture, geography, etc.) would test whether the observed gains extend beyond intra-domain realism. Adaptive data selection strategies could also be explored, where Unity and CIA samples are dynamically sampled according to model uncertainty or feature-space coverage. Finally, integrating domain adaptation objectives such as adversarial feature alignment, or perceptual loss minimization, could mitigate residual distributional drift between mixed domains and the target real domain.

\section{Conclusion}

In conclusion, this study provides a principled framework for quantifying the relative contributions of simulated and generative data, in real-world object detection. The empirical results demonstrate that small, well-calibrated proportions of synthetic or generative data, can significantly enhance generalization. On the other hand, excessive inclusion induces domain drift. Unity data offers geometric variability, CIA offers photometric fidelity, and Real data provides semantic grounding. Their optimal combination maximizes performance without compromising realism. Future work will explore adaptive mixing schedules and cross-domain fine-tuning strategies, to dynamically balance these complementary effects. Grounding dataset composition in quantitative $\Delta$ metrics enables systematic, data-driven evaluation of how synthetic and generative sources influence model robustness and generalization.

\begin{acks}
We thanks INFRABEL for their constant dedication to push research forward.
\end{acks}

\bibliographystyle{ACM-Reference-Format}
\bibliography{sample-base}

\end{document}